# Mpox Detection Advanced: Rapid Epidemic Response Through Synthetic Data

**Yudara Kularathne MD[1], Prathapa Janitha[1], Sithira Ambepitiya MBBS[1],**

**Prarththanan Sothyrajah[1], Thanveer Ahamed[1], Dinuka Wijesundara[1]**

(1) HeHealth Inc. San Francisco, USA

## Abstract

Rapid development of disease detection models using computer vision is crucial in responding to medical emergencies, such as epidemics or bioterrorism events. Traditional data collection methods are often too slow in these scenarios, requiring innovative approaches for quick, reliable model generation from minimal data. Our study introduces a novel approach by constructing a comprehensive computer vision model to detect Mpox lesions using only synthetic data. Initially, these models generated a diverse set of synthetic images representing Mpox lesions on various body parts (face, back, chest, leg, neck, arm) across different skin tones as defined by the Fitzpatrick scale (fair, brown, dark skin). Subsequently, we trained and tested a vision model with this synthetic dataset to evaluate the diffusion models' efficacy in producing high-quality training data and its impact on the vision model's medical image recognition performance. The results were promising; the vision model achieved a 97% accuracy rate, with 96% precision and recall for Mpox cases, and similarly high metrics for normal and other skin disorder cases, demonstrating its ability to correctly identify true positives and minimize false positives. The model achieved an F1-Score of 96% for Mpox cases and 98% for normal and other skin disorders, reflecting a balanced precision-recall relationship, thus ensuring reliability and robustness in its predictions. Our proposed SynthVision methodology indicates the potential to develop accurate computer vision models with minimal data input for future medical emergencies.

***Keywords: Mpox, Disease Detection, Computer Vision, Diffusion models, Synthetic data, Vision Transformers, DreamBooth,***

# 1 Introduction

The dependency on large, extensive datasets significantly impeded the development and training of computer vision models for identifying visual symptoms of new diseases. This necessity for vast amounts of data becomes a substantial obstacle, especially during emergent health crises such as novel endemics, pandemics, or bioterrorism attacks, where acquiring relevant data is inherently challenging. Conventional methods of data collection and compilation are not only slow but often inadequate to keep pace with the swift emergence and identification of new diseases. Delays in developing detection models can lead to widespread disease and increased pressure on healthcare infrastructures. To address this issue, this paper introduces a novel approach, SynthVision, which utilizes diffusion models for the generation of synthetic medical images based on a minimal set of initial guide images.

Traditionally, synthetic image creation involved Generative Adversarial Networks (GANs). For instance, the research by Aljohani & Alharbe (2022)[1] employed Deep Pix2Pix GAN to generate synthetic medical images, affirming the potential of GANs to create realistic and clinically relevant images, a principle that aligns with our SynthVision initiative. However, GANs often face criticism for their limited diversity in outputs and their unstable training processes [2]. Moreover, they require large datasets, which are typically unavailable in medical crisis situations.

Diffusion probabilistic models represent a significant advancement in the field of computer vision, with proven efficacy in generating high-quality images. Research by Khader et al. (2022)[3] showed that these models could produce high-quality medical imaging data, such as MRI and CT scans, and notably enhance the performance of breast segmentation models under conditions of data scarcity [1]. This capability is essential for emerging diseases where data is scarce and poses significant challenges. The images generated via diffusion models are both diverse and of high fidelity, making them highly effective for training robust vision models.

Recent developments in Denoising Diffusion Probabilistic Models (DDPMs), highlighted by Nichol and Dhariwal's work [4], have marked a considerable improvement in the capabilities of diffusion models. Their methods demonstrated efficient image generation processes, yielding high-quality results with optimized computational demands. The process effectively reverses diffusion, converting noise back into meaningful images, which supports our project's objective to maximize output from minimal inputs in vision model development.

Additionally, Ceritli et al. (2023) [5] explored the utility of diffusion models in creating realistic synthetic mixed-type Electronic Health Records (EHRs), showcasing their benefit in producing more accurate synthetic data [5]. This study highlights the flexibility of diffusion models to handle various data types, an essential attribute in healthcare where diverse data is crucial.

In conclusion, this paper aims to showcase the effectiveness of diffusion models in synthesizing images for healthcare, emphasizing their pivotal role in addressing the challenges of sparse data availability. This method promises to enrich the datasets necessary for training effective vision models, thereby improving diagnostic precision and enhancing patient care amidst rapidly evolving medical emergencies.

# 2 Methodology

## Methodology for Generating Images of Mpox lesions Using Personalized Text-to-Image Diffusion Models

In this phase, we employ a customized text-to-image diffusion model, specifically adapted for generating highly accurate and detailed images of Mpox lesions. Our approach involves a meticulous fine-tuning process using the DreamBooth technique to significantly enhance the capabilities of state-of-the-art diffusion models such as SDXL and SD2. This method focuses on using a limited yet highly targeted set of clinically validated images to capture the nuanced characteristics of Mpox, which are crucial for the applications in diagnostics, research, and medical education. All training for the diffusion and diagnostic models was conducted using Google Colab, utilizing NVIDIA A100 GPUs.

## 2.1. Initial Image Collection and Model Personalization:

The initial stage involves the careful curation of eight distinct sets of images, each set containing 15 clinically validated images depicting Mpox lesions. These sets comprehensively cover lesions on various body parts, including the face, back, leg, neck, and arm. We also include images representing three different skin types, classified according to the Fitzpatrick scale: fair skin (types 1 & 2), brown skin (types 3 & 4), and dark skin (types 5 & 6). This diverse collection ensures that the dataset reflects the wide demographic variability of the disease.

Each image is meticulously formatted and annotated with detailed descriptive texts that align the visual content. The text prompts are crafted to include a unique identifier for the Mpox class and describe the clinical appearance of the lesions in explicit detail, enhancing the training process and enabling the model to focus on and reproduce the specific attributes of Mpox lesions accurately.

## 2.2 Synthetic Image Generation and Selection:

From this fine-tuned model, we generated 200 synthetic images for each of the eight validated sets, totaling 1600 images. Subsequently, with clinical guidance, we selected 100 to 150 images from each set, resulting in a highly organized dataset of 1000 synthetic images. In our image generation tasks, we utilized both SDXL and SDv2 models, with fine-tuning via Dreambooth [6]. The SDXL diffusion model, an advanced generative framework, employs stochastic processes within a U-Net architecture to transform noise into high-fidelity images. It offers enhanced sample diversity and robustness over traditional models, despite its higher computational requirements. This positions the SDXL model as a promising candidate for future improvements in generative technology.

## 2.3 Development of the Classification Model

In the development of our classification model, we utilized synthetic data generated by text-to-image diffusion models to construct a comprehensive and heterogeneous training dataset. Our objective was to develop a diagnostic tool capable of accurately distinguishing between mpox-infected skin, normal non-disease skin, and other prevalent dermatological conditions. To achieve this, we combined synthetic images representing mpox with actual patient images of normal non-disease skin and other common skin disorders in our training dataset.

## 2.4 Dataset Preparation

The composition of our training dataset is critical to the model's success. It includes 1000 synthetic images specifically generated to portray Mpox, equipping the model to identify distinctive mpox characteristics. In contrast, the dataset also contains 1000 actual patient images of normal skin and an additional 1000 actual patient imagesof various other skin disorders. These images were sourced directly from within our organization, providing a rich spectrum of real-world conditions that enable the model to refine its ability to distinguish mpox from other skin anomalies effectively. All the sourced images was taken with full patient consent.

Extensive preparations were made in compiling the dataset to ensure the model's robustness and efficacy. Each category, Mpox, normal non-disease skin, and other skin disorders was meticulously balanced with an equal number of images (1000 each), ensuring that the training set is representative of diverse real-world scenarios. This balanced approach helps minimize bias and enhances the model's performance across varied diagnostic contexts.

Model tuning and architectural decisions were primarily informed using a carefully selected validation dataset. This set comprises 150 clinically verified actual patient images for each category (mpox, normal, and other skin disorders). The use of actual patient images in the validation set is important in fine-tuning the model parameters and preventing overfitting, offering a practical test to ensure the model's effectiveness under conditions that closely mimic real-world applications.

The model's performance was evaluated using a test dataset comprised of 100 actual patient images for each category, curated by trained physicians to represent all conceivable variations and severity levels of mpox. This testing set is an essential tool for unbiased performance assessment, simulating real-world scenarios to effectively gauge the model's practical efficacy in clinical settings. This comprehensive evaluation ensures that our classification model is not only theoretically robust but also highly reliable and accurate in practical medical diagnostics.

| Category | Training (Images) | Validation (Images) | Test (Images) |
|---|---|---|---|
| *M-pox* | 1000 (All Synthetic) | 150 (Real) | 100 (Real) |
| *Normal* | 1000 (Real) | 150 (Real) | 100 (Real) |
| *Other* | 1000 (Real) | 150 (Real) | 100 (Real) |

*Table 1: Dataset*

## 2.5 Model Architecture Overview

In our research, we utilize a configuration of the Vision Transformer (ViT) designed specifically for image analysis. This model architecture processes images by dividing them into patches of 16x16 pixels. To better suit the requirements of our study, we adapted the model to accept inputs of 384x384 pixels. After conducting experiments, we found that this higher resolution input significantly improves the model's ability to handle and analyze image data, leading to enhanced accuracy in our image classification tasks.

## 2.6 Enhancements to the ViT Architecture

To optimize our Vision Transformer architecture, we implemented two key modifications: the addition of attention dropout and an extra dense layer with 128 neurons. These enhancements are designed to improve the model's focus during training and increase its capacity to recognize complex image patterns. As a result, our modified architecture achieves superior accuracy in disease classification tasks compared to the standard Vision Transformer setup.

## 2.6 Model Configuration Summary

Our model configuration is finely tuned to extract detailed and localized features by processing images in 16x16 pixel patches. We have optimized the input size from the standard 224x224 pixels to 384x384 pixels, better accommodating the specific needs of our data structure and enhancing the model's performance. The training strategy leverages advanced image preprocessing and augmentation techniques provided by Keras's ImageDataGenerator library. These techniques include rescaling to normalize pixel values, introducing rotational variations, and brightness adjustments to ensure robust performance under varied lighting conditions.

Table 1: Detailed Model Summary

| Layer (type) | Output Shape | Param # |
|---|---|---|
| input_1 (InputLayer) | [(None, 384, 384, 3)] | 0 |
| embedding (Conv2D) | (None, 24, 24, 768) | 590592 |
| reshape (Reshape) | (None, 576, 768) | 0 |
| class_token (ClassToken) | (None, 577, 768) | 768 |
| Transformer/posembed_input (AddPositionEmbs) | (None, 577, 768) | 443136 |
| Transformer/encoderblock_0 (TransformerBlock) | ((None, 577, 768), (None, 12, None, None)) | 7087872 |
| Transformer/encoderblock_1 (TransformerBlock) | ((None, 577, 768), (None, 12, None, None)) | 7087872 |
| ⋮ | ⋮ | ⋮ |
| Transformer/encoderblock_11 (TransformerBlock) | ((None, 577, 768), (None, 12, None, None)) | 7087872 |
| Transformer/encoder_norm (LayerNormalization) | (None, 577, 768) | 1536 |
| ExtractToken (Lambda) | (None, 768) | 0 |
| flatten (Flatten) | (None, 768) | 0 |
| dense (Dense) | (None, 128) | 98432 |
| dense_1 (Dense) | (None, 2) | 258 |
| **Total params:** | | 86090496 |
| **Trainable params:** | | 86090496 |
| **Non-trainable params:** | | 0 |

*Figure 1: Model Architecture*

## 2.7 Model Parameters and Hyper Parameter Tuning

The development phase involved a hyperparameter tuning process aimed at optimizing the model's performance. We conducted extensive experiments with various batch sizes, learning rates, and image dimensions. The optimal settings were identified as a batch size of 32, an image size of 384x384 pixels, and a learning rate of 1e-4. To enhance accuracy, we implemented a dynamic reduction of the learning rate when no further improvements were observed. Additionally, we used the Adam optimizer to facilitate efficient convergence and applied early stopping mechanisms to prevent overfitting, thereby ensuring the model's robustness.

# 3. Results

## 3.1 Generated Synthetic Data

Below, displays generated synthetic data from different body parts (Face, Back, Leg, Neck, Arm) and different skin types according to the Fitzpatrick scale. In the Fitzpatrick scale, there are 6 skin types. We have divided these six into 3 categories: fair skin (types 1 & 2), brown skin (types 3 & 4), and dark skin (types 5 & 6).

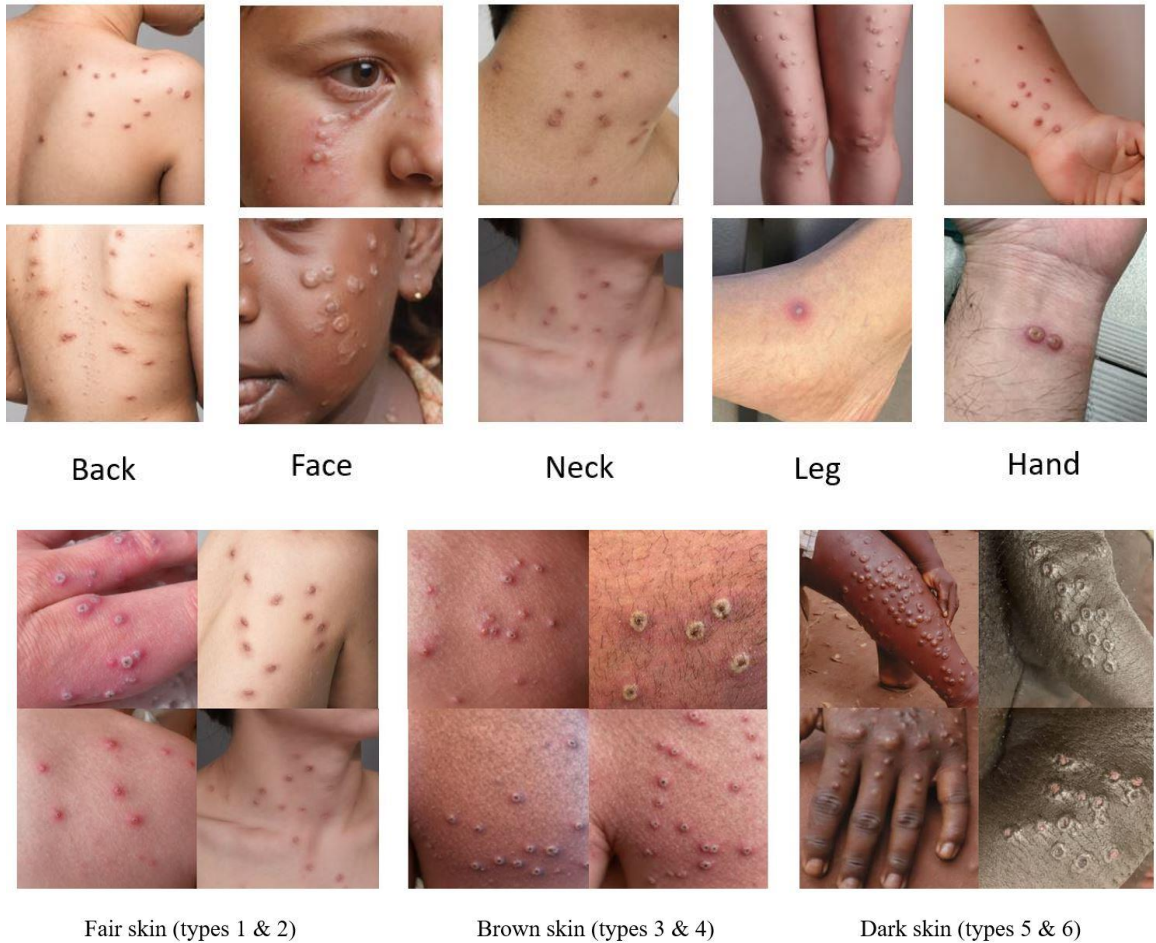

Figure 2 : Sample Generated Images

## 3.2 Classification Results

The evaluation of the classification model on the designated test set has yielded highly encouraging results, as shown by the confusion matrix and the detailed classification report. The confusion matrix reveals a substantial number of correct predictions across the three defined categories M-pox, Normal, and Other with few instances of misclassification.

The model achieved exceptional accuracy in identifying Normal conditions, with 98 true positives out of 100 instances, reflected in a recall of 0.98 and a precision of 0.97, resulting in an F1-score of 0.98. This indicates a significant capability of the model to reliably identify and categorize normal skin conditions without error, essential for reducing false-positive rates in clinical applications.

For the M-pox category, the model recorded 96 true positives, maintaining a high precision and recall of 0.96, culminating in an F1-score of 0.96. This demonstrates an enhanced specificity in detecting M-pox cases, crucial for targeted medical interventions and treatments. Similarly, the other category displayed comparable performance with 96 true positives, a precision of 0.97, and a recall of 0.96, achieving an F1-score of 0.96, confirming the model's effectiveness in accurately classifying diverse and less common skin conditions.

Overall, the model's aggregated performance metrics from the classification report—precision, recall, and F1-score—all stand impressively at 0.97 across a total of 300 instances. These metrics not only reflect the model's accuracy but also its consistency in performance across different categories. The accuracy of 0.97 for the entire model showcases its exceptional overall effectiveness and reliability in a clinical setting, making it an invaluable tool for the diagnosis and classification of various skin conditions. The consistency in high performance across categories ensures that the model can be reliably used in diverse diagnostic scenarios, maintaining high standards of care and accuracy.

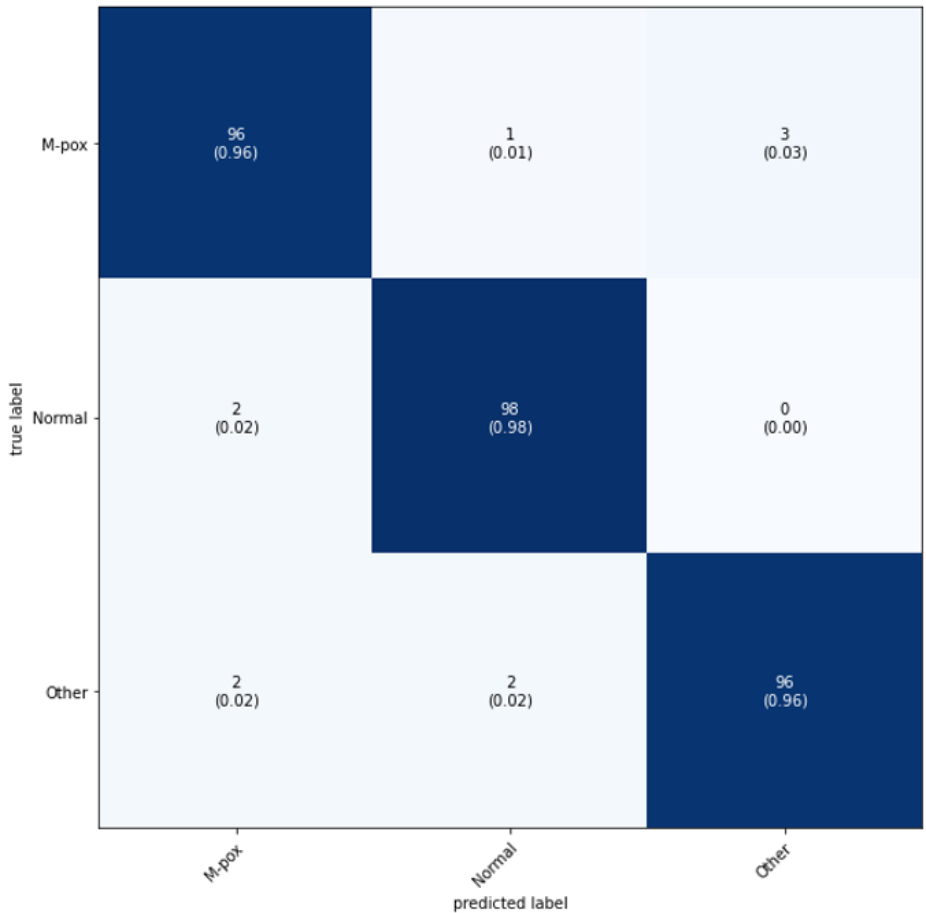

*Figure 3 : Confusion Matrix*

```
Classification Report:
              precision    recall  f1-score   support

       M-pox       0.96      0.96      0.96       100
      Normal       0.97      0.98      0.98       100
       Other       0.97      0.96      0.96       100

    accuracy                           0.97       300
   macro avg       0.97      0.97      0.97       300
weighted avg       0.97      0.97      0.97       300
```

*Figure 4 : Classification Report*

## 4. Conclusion

In conclusion, our study demonstrates the transformative impact of synthetic data on the development of medical diagnostic models, particularly in scenarios demanding rapid and accurate disease detection. Through SynthVision approach, leveraging advanced diffusion models to generate synthetic images, we achieved remarkable diagnostic accuracy of 97% - precision, recall, and F1-scores consistently above 0.96 in identifying Mpox, normal, and other skin conditions. These results affirm the capability of synthetic data to not only enhance model training but also ensure the model's reliability in clinical settings, paving the way for its adoption in urgent healthcare responses where traditional data acquisition might be too slow. This pioneering approach sets a new benchmark for deploying artificial intelligence effectively in the fight against epidemics and other medical emergencies.